\documentclass[10pt,journal]{IEEEtran}
\usepackage{array}
\usepackage{float}
\usepackage{url}
\usepackage{multirow}
\usepackage{amsmath}
\usepackage{amssymb}
\usepackage{cancel}
\usepackage{graphicx}
\usepackage{setspace}
\usepackage{cite}
\usepackage{verbatim}
\usepackage{makecell}
\graphicspath{{figures/}}

\makeatletter

\providecommand{\tabularnewline}{\\}
\floatstyle{ruled}
\newfloat{algorithm}{tbp}{loa}
\providecommand{\algorithmname}{Algorithm}
\floatname{algorithm}{\protect\algorithmname}

\ifCLASSINFOpdf
\else
\fi
\begin{document}
%
\title{OTFace: Hard Samples Guided Optimal Transport Loss for Deep Face Representation}
%
%
%
%


\author{\IEEEauthorblockN{Jianjun Qian, Shumin Zhu, Chaoyu Zhao, Jian Yang, and Wai Keung Wong }
	
\IEEEcompsocitemizethanks{\IEEEcompsocthanksitem Jianjun Qian, Shumin Zhu, Chaoyu Zhao and Jian Yang are with the PCA Lab, Key Lab of Intelligent Perception and Systems for High-Dimensional Information of Ministry of Education, and Jiangsu Key Lab of Image and Video Understanding for Social Security, School of Computer Science and Engineering, Nanjing University of Science and Technology, Nanjing, Jiangsu,China.  (e-mail: csjqian@njust.edu.cn, zhushumin@njust.edu.cn, cyzhao@njust.edu.cn, csjyang@njust.edu.cn).
	
\IEEEcompsocthanksitem Wai Keung Wong is with the Institute of Textiles and Clothing at The Hong Kong Polytechnic University, Hong Kong. (e-mail: calvin.wong@polyu.edu.hk).} }
\IEEEtitleabstractindextext{%
\begin{abstract}
Face representation in the wild is extremely hard due to the large scale face variations. To this end, some deep convolutional neural networks (CNNs) have been developed to learn discriminative feature by designing properly margin-based losses, which perform well on easy samples but fail on hard samples. Based on this, some methods mainly adjust the weights of hard samples in training stage to improve the feature discrimination. However, these methods overlook the feature distribution property which may lead to better results since the miss-classified hard samples may be corrected by using the distribution metric. 
This paper proposes the hard samples guided optimal transport (OT) loss for deep face representation, OTFace for short. OTFace aims to enhance the performance of hard samples by introducing the feature distribution discrepancy while maintain the performance on easy samples. Specifically, we embrace triplet scheme to indicate hard sample groups in one mini-batch during training. OT is then used to characterize the distribution differences of features from the high level convolutional layer. Finally, we integrate the margin-based-softmax (e.g. ArcFace or AM-Softmax) and OT to guide deep CNN learning. Extensive experiments are conducted on several benchmark databases. The quantitative results demonstrate the advantages of the proposed OTFace over state-of-the-art methods.
\end{abstract}

}

\maketitle

\IEEEdisplaynontitleabstractindextext

%
\IEEEpeerreviewmaketitle


%
%
%
%
\section{Introduction}\label{sec:introduction}
\IEEEPARstart{F}{ace}
recognition (FR) has been an active topic in the field of pattern recognition and computer vision. In the past decades, numerous of methods have been developed by world-wide researchers. Among these methods, many well-known face representation models have been widely used in practical applications such as biometric identification, video surveillance and human-computer interaction. 

The holistic approach of face representation motivates the low-dimension subspace representation methods, such as linear subspace \cite{97-Eigenfaces}, manifold learning \cite{05-LaplacianFace,07-UDP}, sparse representation \cite{09-SRC,13SparseDP} and low rank \cite{11-RPCA,17-NMR}.  These methods achieve better results in constrained  environments but fail to address the face representation in un-constrained environments. Thus, local feature representation methods are proposed to solve robust face recognition task.  Gabor feature is proven to be robust to illumination and facial expression by capturing desirable local characteristic structure \cite{02-LiuGabor}. Local Binary Pattern (LBP) employed the differences between central pixel and its neighbors over the local patch to characterize the local structure \cite{06-LBP}.  Some works combined Gabor and LBP, as well as their multi-level extensions, performed better results in FR \cite{10-LBPGabor,13-BlessingHD,18-CompressiveBinary}. 
It is pity that handcrafted feature is still lack of distinctiveness and compactness in dealing with real-world FR tasks. Subsequently, learning-based local feature methods were developed in FR area.  Lei et al. presented the learning discriminant face descriptor in a data-driven way and learn the discriminative local features to minimize the distance between the same person and maximize that between different persons \cite{14-DFD}.  From the view of LBP, compact  binary face descriptor feature learning is proposed for robust face representation \cite{15-CBFD}. However, the above mentioned methods can be considered as shallow face representation methods which still have limitations to address the complex nonlinear face variations in practical applications. 

\begin{figure*} \label{fig:OTFaceframe}
	\noindent 
	\begin{centering}
		\includegraphics[scale=0.6]{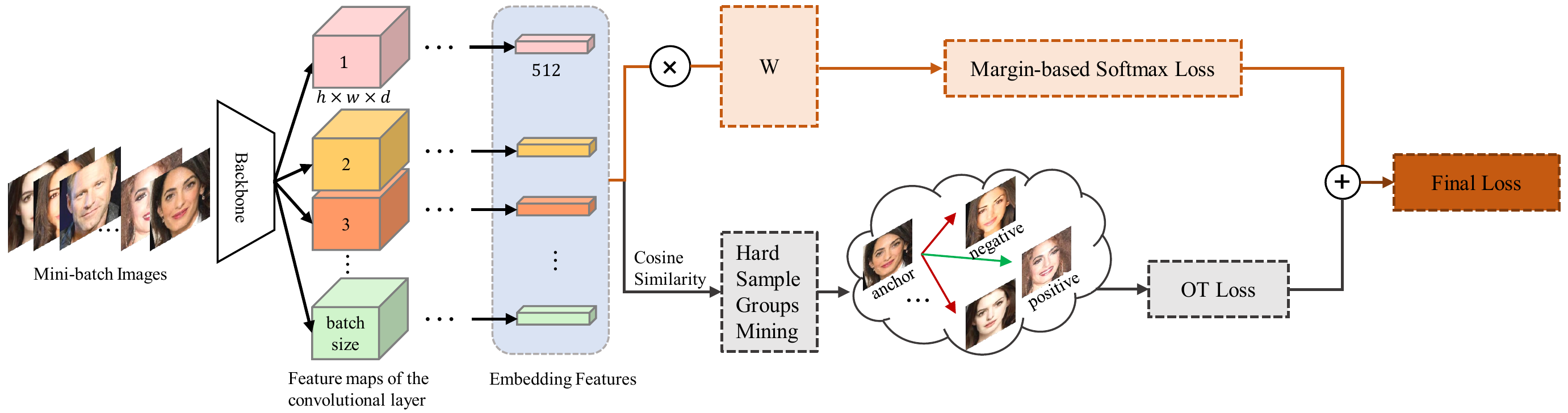}
		\par\end{centering}
	\begin{singlespace}
		\caption{The pipeline of the proposed OTFace.}
	\end{singlespace}
\end{figure*}

The advent of deep learning has changed this situation. Specifically, AlexNet achieved the best performance in ImageNet competition which sets off a wave of deep learning based methods in FR. Actually, deep convolutional neural network characterizes the multi-level representation with different granularities by employing the cascade of neural network processing unit. Subsequently, the CNN architectures, such as AlexNet \cite{12AlexNet}, VGGNet \cite{14-VGGNet}, GoogleNet \cite{15-GoogleNet} and ResNet \cite{16-ResNet}, are widely used to characterize the facial feature. Taigman et al. derived a nine-layer deep neural network to represent facial feature in conjunction with an efficient face alignment step (DeepFace) for robust FR \cite{14-DeepFace}. Based on DeepFace, DeepID series \cite{14-DeepID,14-DeepID2,15-DeepID3} aim to learn the discriminative face representation through the large scale face identification, joint face identification-verification and very deep neural network structures. In \cite{15-FaceNet}, the authors presented a face recognition system named FaceNet, which combined the deep convolutional network and triplet loss to obtain the low-dimensional and compact facial feature vector. Parkhi et al. provided a very large scale dataset and used VGGNet to achieve comparable performance in accuracy of face recognition \cite{15-VGGFace}. In addition, there are still numbers of novel network architectures are designed for FR to improve the efficiency and robustness. 
Li et al. developed a novel deep locality-preserving constitutional neural network to improve the discriminative power of deep feature \cite{19-DLP-CNN}.
Choi et al. employed the  Gabor feature maps as the input of Deep CNNs and then combined the multiple outputs of Deep CNNs for feature ensemble \cite{20-GaborDCNN}.
Zhang et al. proposed a deep cascade model with hierarchical learning for corrupted FR \cite{20-DCM}. In , the authors aims to improve the performance in real-world FR by proposing the SFace to optimize intra-class and inter-class distance to some extent \cite{21-TIP-SFace}. 

From the view of loss function, Softmax based methods can obtain remarkable results based on large scale training data and properly designed DCNN architectures. However, for FR, the intra-variations could be larger than inter-difference caused by occlusion, illumination changes, pose and expression variations. Therefore, the learned features are not discriminative enough by softmax loss. The triplet loss is used to increase the margin for enhancing the discriminative power of learned feature \cite{14-DeepFace,15-FaceNet}. The fly in the ointment is training instability of triplet loss due to the selection of effective pair-wise training samples. To encourage the intra-class compactness, Wen et al. developed a center loss to learn the distance between each feature vector and its class center \cite{16-CenterLoss}. In \cite{16-LSoftmax}, lagre margin softmax (L-softmax) added angular constraints to improve the feature discrimination. Subsequently, SphereFace is proposed to learn discriminative facial feature by increasing the inter-class margin and shrinking the intra-class angular distribution \cite{17-SphereFace}.  AM-Softmax and CosFace directly used cosine margin penalty to further encourage the decision margin in angular space \cite{18-CosFace,18-AMSoftmax}. Deng et al. proposed an additive angular margin loss (ArcFace) to improve the discriminative power of deep facial feature, which achieved better performance on various large-scale face image/video datasets \cite{19-ArcFace}. 

Although the above methods have achieved remarkable results, they ignore the role of hard (mis-classified) samples for learning discriminative feature. To
overcome this problem, Liu et al. proposed the Adaptive Margin Softmax and introduced the Hard Prototype Mining strategy to concentrate on hard classes during classification training \cite{19-AdativeFace}. In \cite{16-HMSoftmax}, the authors directly ignored the easy samples to concentrate on hard samples in training stage. Focal loss based softmax is then introduced to re-weight all samples and made hard samples have large loss \cite{17-FocalLoss}.  
Wang et al. combined the advantages of feature margin and feature mining strategy into a unified framework for deep face recognition (MV-Softmax) \cite{19-MVSoftmax}. Specifically, they employed the feature vectors of hard samples to guide the discriminative feature learning. Different from MV-Softmax, CurricularFace automatically learns from the easier samples first and harder samples later \cite{20-CurricularFace}. Moreover, it also adaptively re-weights hard samples in different training stages. However, these methods mainly adjust weights of samples and make the harder ones to have larger loss values. They overlook the embedding feature distribution information of samples. In other words,  some mis-classified samples in cosine space may be corrected from the view of distribution metric. Based on this point, Optimal Transport (OT) is brought to our attention. This is because OT is an efficient tool to characterize the distribution differences between samples compared with Kullback-Leibler divergence and Euclidean distance. Thus, Wasserstein discriminative analysis employed the OT to capture the interactions of global distribution and local structure \cite{Flamary2018}. In \cite{OTdomain}, they developed the OT regularized framework to solve the domain adaption problem. Based on the distribution metric merits of OT,  Qian et al. integrated nuclear norm and optimal transport into a unified model to solve occlusion problem of image classification \cite{20-OTR}. 
Additionally, a number of methods have been proposed based on OT to address pattern recognition problem \cite{pmlr-v51-rolet16,18-WassersteinDA}. 

Motivated by the above mentioned methods,  we develop a hard samples guided scheme to improve the performance of face representation, which highlights the role of hard samples in the training stage to learn discriminative feature by using the OT loss from the distribution scale. The pipeline of the proposed method is shown in Fig. 1.  To sum up, the main contributions of this paper can be summarized as follows : 

\begin{itemize}
	\item We borrow the idea of triplet loss to indicate the hard sample groups, and then employ these hard sample groups to enhance the discriminative feature learning.
	\item OT is introduced to characterize the differences of feature maps from the distribution scale. We combine the OT loss and margin-based-softmax to formulate the loss function for robust deep face representation.
	\item We conduct experiments on eight benchmarks, including the LFW, AgeDB, CALFW, CPLFW, CFP, IJB-B, IJB-C and MegaFace. Experimental results demonstrate the advantages of the proposed methods over the state-of-the-art deep face representation methods.
\end{itemize}

The remainder of this paper is organized as follows: Section II introduces the proposed OTFace, including hard sample groups mining and optimal transport loss. Section III provides more analysis between our OTFace and the related Softmax based methods. Section IV conducts experiments on several benchmarks to demonstrate the advantages of the proposed methods. We conclude the paper in Section V.

\section{The Proposed OTFace}
This section mainly introduces the OTFace, a novel hard samples guided scheme for learning discriminative feature for deep FR. The proposed OTFace consists of two parts. The goal of the first part is to explore the hard sample groups based on the idea of triplet loss. In the second part, OT is introduced to describe the feature distribution discrepancy between samples. Then OT loss and margin-based-softmax (AM-Softmax or ArcFace) are integrated together to construct the loss function for robust FR. 

\begin{figure}
	\noindent \begin{centering}
		\includegraphics[scale=0.6]{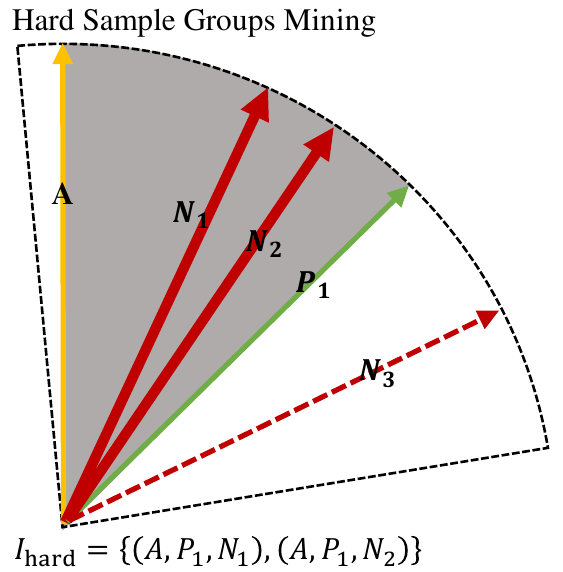}
		\par\end{centering}
	\begin{singlespace}
		\caption{Hard sample groups mining. A denotes the current anchor sample, $ P_1 $ denotes the positive sample, $ N_1, N_2 and N_3 $ are negative samples. If the negative samples lie in the space (gray area) between anchor (A) and positive sample ($ P_1 $) , the corresponding sample groups are considered as hard sample groups. }
	\end{singlespace}
\end{figure} \label{fig:Hardsample}

\subsection{Hard Sample Groups Mining}\label{section:Hardmining}
It is a naive idea to improve the face representation results by hard samples mining. The previous works employed different strategies to indicate hard samples and adjusted weights of them to pay more attention in training stage. Different from these work, we employ the triplet loss to indicate hard sample groups. Next, this section introduce the details of how we explore the hard sample groups during training. 

Given $ N $ samples in a mini-batch, let each sample $ x_i (i=1,2, \cdots, N) $ as the anchor in order, and compute the cosine distances between anchor sample and the other samples. Then, we construct a series of  sample groups, including the anchor sample, the positive sample is from the same class with the anchor, and the negative one comes from the different class with the anchor. Based on this, the sample groups are considered as hard sample groups should satisfy the following rule: the cosine distance between the anchor and the positive sample is smaller than that between the anchor and the negative sample as shown in Fig. 2. The formulation is as follows:

\begin{equation}
consine ( f(x_i^a) ,  f(x_j^p)  )  < consine ( f(x_i^a) ,  f(x_k^n)  )
\end{equation}
where $ f(x_i ) \in \Re^{d} $ represents the embedding feature of $ i $-th sample. $ x_i^a $ means the $ i $-th sample is anchor.  $ x_j^p $ means the $ j $-th sample is positive sample, $ x_k^n $ means the $ k $-th sample is negative sample. 

In our opinion, the hard sample groups with large scale variations can satisfy the above rule since the cosine distances of within-class samples are larger than that of between-class samples in general. The detail of our hard sample group mining scheme is shown in Algorithm 1.

\begin{algorithm}
	\caption{Hard sample groups mining\label{alg:Hardsample}}
	
	\textbf{Input: } Embedding features of sample images in a mini-batch $ [ f(x_1), f(x_2), \cdots, f(x_N)  ]  $, and the corresponding labels.
	
	Initialize the hard sample group set $ I_{hard} = \varnothing $;
	
	for $ i = 1 : N $
	
	\qquad $ a = f(x_i) $        
	
	\qquad select the positive samples of $ a $ to construct set $ \boldsymbol{pos} $;
	
	\qquad select the negative samples of $ a $ to construct set $ \boldsymbol{neg} $;
	
	\qquad for $ j = 1 : length(\boldsymbol{pos}) $
	
	\qquad \qquad  $ a_p = \boldsymbol{pos}(j); $
	
	\qquad \qquad for $ k = 1 : length(\boldsymbol{neg}) $
	
	\qquad \qquad \qquad $ a_n = \boldsymbol{neg}(k); $
	
	\qquad \qquad \qquad if $  (a, a_p, a_n) $ satisfies the Eq.(1)
	
	\qquad \qquad \qquad then $  (a, a_p, a_n) $ belongs to $ I_{hard} $;
	
	\qquad \qquad end
	
	\qquad end
	
	end
	
	\textbf{Output}: Hard sample groups $ I_{hard} $.
\end{algorithm}

\subsection{Optimal Transport Loss}

\subsubsection{Outline of Optimal Transport}
In this subsection, we briefly review the optimal transport minimization problem on discrete distribution. Given two discrete samples $ \boldsymbol{a} \in \Re^n $ and $ \boldsymbol{b} \in \Re^m $,  the empirical distributions of them are:

\begin{equation}
\boldsymbol{a} =  \sum_{i=1 }^{n} p_i^{a} \delta_{\boldsymbol{a}_i},  \space  \boldsymbol{b}= \sum_{i=1 }^{m} p_i^{b} \delta_{\boldsymbol{b}_i}
\end{equation}
where $ \delta_{\boldsymbol{a}_i} $ is the Dirac at position $ \boldsymbol{a}_i $ and $ \delta_{\boldsymbol{b}_i} $ is the Dirac at position $ \boldsymbol{b}_i $. $ p_i^a $ is the probability associated to the $ i $-th element of sample $ \boldsymbol{a} $. 
Based on Kantorovich relaxation, the set of probabilistic of couplings linking a pair of distributions $ (\boldsymbol{a}, \boldsymbol{b}) $ is defined as \cite{18-WassersteinDA}:

\begin{equation}
\boldsymbol{\rm U}(\boldsymbol{a}, \boldsymbol{b}) = \{  \boldsymbol{\rm P }\in \Re^{n \times m} |  \boldsymbol{\rm P} \boldsymbol{\rm 1}_m = \boldsymbol{1}/n,   \boldsymbol{\rm P} \boldsymbol{\rm 1}_n = \boldsymbol{1}/m  \}
\end{equation}
where $ \boldsymbol{\rm P}_{i,j} $ describes the amount of mass flowing from element $ i $ to element $ j $. $ \boldsymbol{\rm 1}_n $ is a $ n $-dimensional vector of ones. $ \boldsymbol{\rm 1}_m $ is a $ m $-dimensional vector of ones. The optimal transport with Kantorovitch relaxation is:

\begin{equation}
{\rm OT}(\boldsymbol{a}, \boldsymbol{b}) = \underset{\boldsymbol{\rm P} \in \boldsymbol{\rm U}(\boldsymbol{a}, \boldsymbol{b})}{min} <\boldsymbol{\rm C}, \boldsymbol{\rm P}> = \sum_{i,j} \boldsymbol{\rm C}_{i,j} \boldsymbol{\rm P}_{i,j}
\end{equation} 
where $ \boldsymbol{\rm C} $ is the cost matrix and $ \boldsymbol{\rm C}_{i,j} $ is the cost from $ a_i $ to $ b_j $. 
$ <\cdot , \cdot> $ represents the Frobenius dot product. Eq. (4) can be solved by the standard linear program algorithm and its solutions are not necessarily unique. Additionally, this model spend more computation time when facing high dimensional histograms. To address these problems,  the entropy regularized OT is proposed to approximate the original OT solution \cite{13Sinkhorn} :
\begin{equation}
	{\rm OT}(\boldsymbol{a}, \boldsymbol{b}) = \underset{\boldsymbol{\rm P} \in \boldsymbol{\rm U}(\boldsymbol{a}, \boldsymbol{b})} {min}
	<\boldsymbol{\rm P}, \boldsymbol{\rm C}> - \varepsilon \boldsymbol{H} (\boldsymbol{\rm P}) 
\end{equation}
where $ \boldsymbol{H} (\boldsymbol{\rm P}) = -\sum_{i,j} \boldsymbol{\rm P}_{i,j} ( log(\boldsymbol{\rm P}_{i,j}) - 1 ) $ is the discrete entropy of transport matrix. The solution to (5) is unique and has the form $ \boldsymbol{\rm P}_{i,j} = \boldsymbol{u}_i \boldsymbol{\rm K}_{i,j} \boldsymbol{v}_j $, where $ \boldsymbol{\rm K}_{i,j} = e^{-\eta \boldsymbol{\rm C}_{i,j}} $. For unknown scaling variables $ \boldsymbol{u} $ and $ \boldsymbol{v} $, they can be solved by using the iterative scheme Sinkhorn's algorithm \cite{13Sinkhorn}.

\subsubsection{OT based Loss Function}
Intuition says that the well-separated (easy) samples have little effect on discriminative feature learning. In other words,  the hard sample groups mined in the previous subsection are more important to improve the feature discriminablity. However, it is difficult to enhance the discriminative power of deep CNN feature by just using cosine (or Euclidean) distance as loss constraint. OT is the efficient tool to characterize the distribution differences, which motivates us to introduce the OT into the loss constraint from the distribution scale. 
Although it is straightforward to adapt embedding feature for computing OT loss, it is not a good choice. This is because embedding feature cannot reveal the intrinsic distribution property of 2D image. To overcome this issue, we employ the feature maps of the convolution layer to compute the distribution differences between two image samples. 

Given two facial images $ \boldsymbol{x}_1 $ and $ \boldsymbol{x}_2 $, deep CNN is employed to represent the facial feature.  Here, the feature maps of the convolution layer can be considered as a tensor structure feature from the geometric perspective as shown in Fig. 3. As well known, each feature map is achieved by using the different convolution kernels, which reveal the intrinsic feature of facial image from different view. From Fig. 3, we can see that there is a $ d $-dimension vector for each pixel of the frontal slice if the tensor structure feature is composed of $ d $ feature maps. 

\begin{figure*}
	\noindent \begin{centering}
		\includegraphics[scale=0.8]{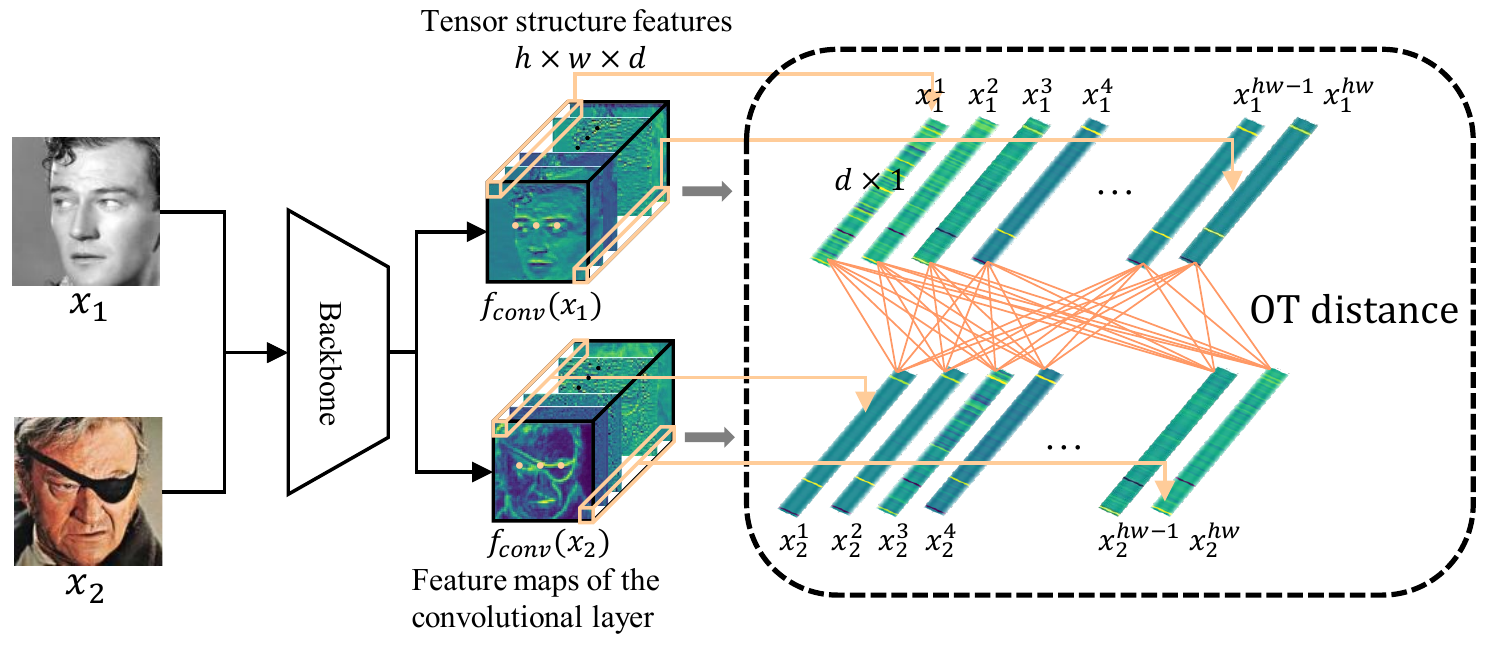}
		\par\end{centering}
	\begin{singlespace}
		\caption{The overview of OT distance between two facial images. }
	\end{singlespace}
\end{figure*} \label{fig:OTLoss}

In this way, the tensor structure feature can be seen as the multivariate distribution since each pixel corresponds to a vector. Specifically, $ f_{Conv}(\boldsymbol{x}_1) \in \Re^{h \times w \times d} $ and  $ f_{Conv}(\boldsymbol{x}_2) \in \Re^{h \times w \times d} $ represent the tensor structure features of $ \boldsymbol{x}_1 $ and $ \boldsymbol{ x}_2 $, respectively. To compute the distribution differences, the tensor structure feature is reformulated as the bivariate distribution $ M(\boldsymbol{x}_1) = [ \boldsymbol{m_1^1},  \boldsymbol{m_1^2}, \cdots, \boldsymbol{m_1^{n}}  ] \in \Re^{n \times d } $ and $ M(\boldsymbol{x}_2) = [ \boldsymbol{m_2^1},  \boldsymbol{m_2^2}, \cdots, \boldsymbol{m_2^{n}}  ] \in \Re^{n \times d } $ ($ n = h \times w $).  The OT distance between $ \boldsymbol{x}_1 $ and $ \boldsymbol{x}_2 $ is defined as:

\begin{equation}
{\rm OT}(M(\boldsymbol{x}_1), M(\boldsymbol{x}_2)) = \underset{\boldsymbol{\rm P} \in \boldsymbol{\rm U}}{min}<\boldsymbol{\rm P}, \boldsymbol{\rm C}> - \varepsilon \boldsymbol{H} (\boldsymbol{\rm P}) 
\end{equation}
where $ \boldsymbol{\rm P} $ is the transport matrix between two samples. $ \boldsymbol{\rm C} $ is cost matrix and has entries $ \boldsymbol{\rm C}_{i,j} = d(\boldsymbol{m_1^i}, \boldsymbol{m_2^j}) $ , where $ d(\cdot,\cdot) $ is the Cosine distance. The set $ \boldsymbol{\rm U} = \{ \boldsymbol{\rm P} \in \Re^{n \times n} |  \boldsymbol{\rm P} \boldsymbol{1}_{n} = \boldsymbol{1}_{n} / n,  \boldsymbol{\rm P}^T \boldsymbol{1}_{n} = \boldsymbol{1}_{n} / n \} $ contains all possible transport matrices . 
Here, we use Sinkhorn algorithm to achieve the solution of problem (6) since Sinkhorn algorithm can speed the computation time and have a unique solution. The solution $ \boldsymbol{\rm P} $ can be expressed as :  

\begin{equation}
\boldsymbol{\rm P} = diag(\boldsymbol{u}) \boldsymbol{\rm K} diag(\boldsymbol{v})
\end{equation}
where $ \boldsymbol{\rm K} = e^{- \varepsilon \boldsymbol{\rm C}} $. Both $ \boldsymbol{u} $ and $ \boldsymbol{v} $ are non-negative vectors of $ \Re^{n} $. Based on this, the transport matrix can be computed by using the fixed-point iterative strategy.  The scaling vectors $ \boldsymbol{u} $ and $ \boldsymbol{v} $ are updated for iteration $ t + 1 $: 

\begin{equation}
\boldsymbol{v}^{t+1} = \dfrac{\boldsymbol{1}_n / n}{K^T\boldsymbol{u}^{t}} , \space  \boldsymbol{u}^{t+1} = \dfrac{\boldsymbol{1}_n / n }{K \boldsymbol{v}^{t+1}}
\end{equation}
where the initialization of $ \boldsymbol{u}^{0} $ is set to $ \boldsymbol{1}_n $.

The previous OT distance metric motivates us to present the OT based loss function, which is defined as follow:

\begin{equation}
\begin{aligned}
\boldsymbol{L}_{OT} =  \sum_{i=1}^{H} & [  {\rm OT} ( \boldsymbol{M}(x^{anchor}_i), \boldsymbol{M}(x^{positive}_i) ) -  \\
& {\rm OT} ( \boldsymbol{M}(x^{anchor}_i), \boldsymbol{M}(x^{negative}_i) ) ]_{+}
\end{aligned}
\end{equation}
where $ H $ represents the number of hard sample groups which are explored by using Algorithm \ref{alg:Hardsample}. $ \boldsymbol{x}^{anchor}_i $ is the anchor of the $ i $-th hard group,  $ \boldsymbol{x}^{positive}_i $ is the positive sample of the $ i $-th hard group and $ \boldsymbol{x}^{negative}_i $ is the negative sample of the $ i $-th hard group.
Finally, we further combine the loss function of ArcFace to maintain the results on easy samples. The final loss function is :

\begin{equation}
\boldsymbol{L}(\theta) = \boldsymbol{L}_{OT} + \boldsymbol{L}_{ArcFace}
\end{equation}
where $ \theta $ represents the parameter set. Here, $ \boldsymbol{L}_{ArcFace} $ can be replaced by $ \boldsymbol{L}_{AM-Softmax} $ or any of the softmax based losses. In our study, we just use two representative margin-based-softmax methods AM-Softmax and ArcFace to maintain the performance of easy samples. 

Actually, the proposed OT loss can also be optimized by the typical stochastic gradient descent (SGD) algorithm, more details see reference \cite{18-SinkhornAutoDiff}. And the whole pipeline of our OTFace is summarized in Algorithm 2. 

\begin{algorithm}
	\caption{OTFace \label{alg:OTFace}}
	
	\textbf{Input: }Training set of images $ S $, training epochs $ \tau $.
	
	\textbf{Initialize}
	$ \alpha = 1 $; The convolutional network parameter $ \theta $ and the last fully-connected layer parameters $ W $ are initialized randomly.
	
	
	\textbf{while} $ \alpha < \tau $  \textbf{do}
	
	\qquad Randomly sample facial images to fetch mini-batch;
	
	\qquad \textit{Forward}: 
	
	\qquad (1) Compute the AM-Softmax (ArcFace) loss; 
	
	\qquad (2) Compute the OT loss according to the hard sample 
	
	\qquad groups.
	
	\qquad \textit{Backward}: 
	
	\qquad Update the parameters $ \theta $ and $ W $.
	
	\textbf{end}
	
	\textbf{Output}: Parameters $ \theta $  and $ W $.
\end{algorithm}

\begin{figure}
	\noindent \begin{centering}
		\includegraphics[scale=0.4]{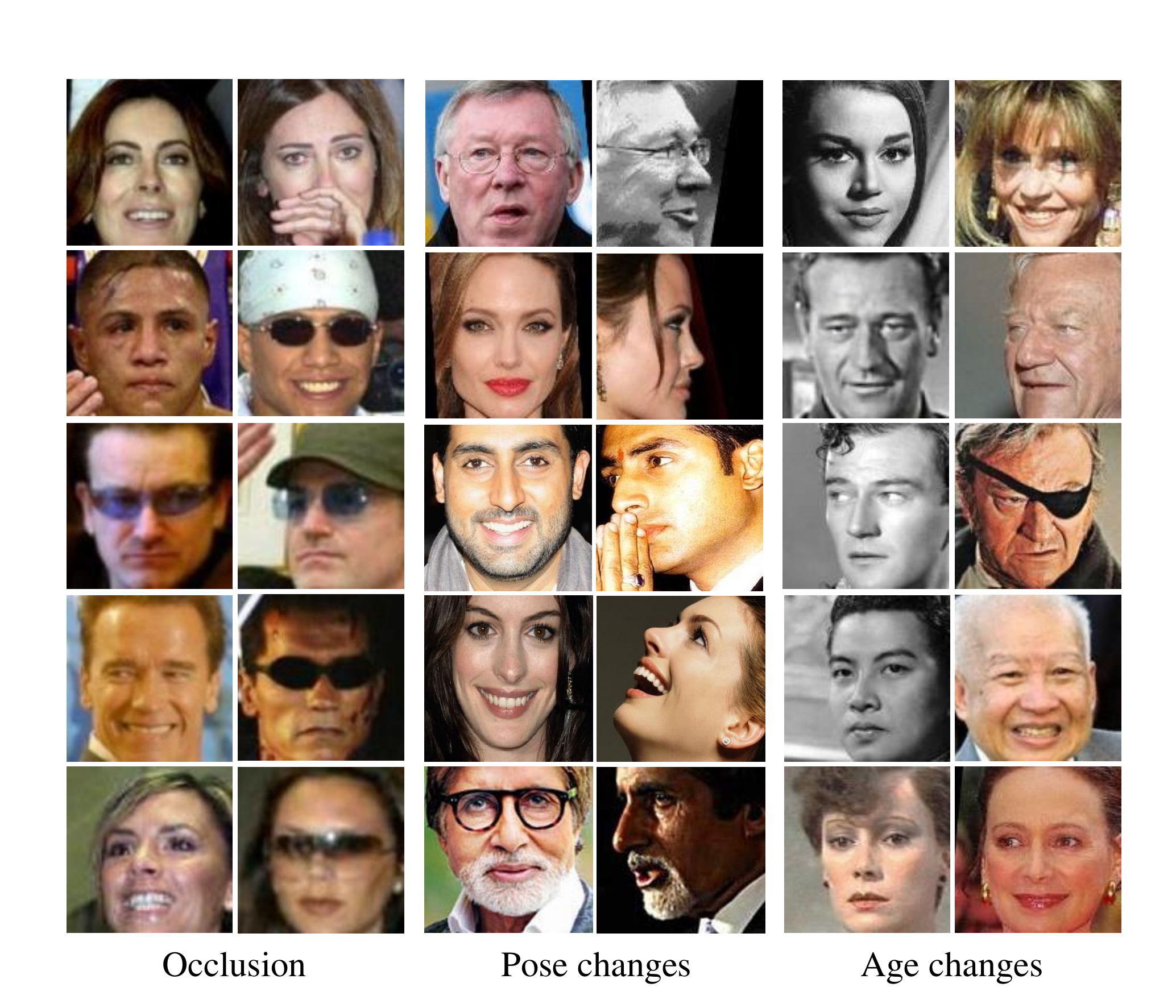}
		\par\end{centering}
	\begin{singlespace}
		\caption{Some face image pairs are mis-classified by using AM-Softmax and are corrected by using OTFace. }
	\end{singlespace}
\end{figure} \label{fig:AMOT}

\section{Further Analysis}

\subsection{Why OTFace}
In this subsection, we provide more analysis to explain the advantages of our OTFace. It is known that the most softmax based methods achieve the better performance on easy samples, while these methods perform poorly on hard samples (e.g face images with occlusions, pose changes and age variations ). In Fig. 4, we select some representative hard sample pairs from various face datasets.  Further, deep CNN with margin based Softmax (in the case of AM-Softmax) constraint leads to wrong results for these facial image pairs. However, the proposed OTFace can give right decisions when dealing with these hard facial images. The possible reasons are that AM-Softmax introduces the adaptive margin into the loss function with feature and weight normalized. It ignores the distribution property to enhance the discriminative power of deep CNN feature. Compared with AM-Softmax, OTFace employs the OT to characterize the sample difference from the distribution scale and maintains the AM-Softmax constraint.

From the view of metric scheme, AM-Softmax computes the cosine loss of  deep embedding feature and ArcFace improve the discriminative power of feature by computing the additive angular loss. For OT loss, we evaluate the distribution differences by using the feature maps of the high level convolution layer. In our opinions, these feature maps contain various high level semantic feature and preserve the structure property to some extent. Based on this, we consider the feature maps as the multi-variant distribution and use OT to explore the intrinsic difference from the distribution scale. 

\begin{table}
	\caption{\label{tab:comparisons} The comparisons between the proposed OTFace with different Softmax based methods.}
	\begin{spacing}{0.9}
		\noindent \centering{}{\small{}}%
		\setlength{\tabcolsep}{1mm}{
			\begin{tabular}{l | l | l }
				\hline
				{\scriptsize{} \textbf{Methods} } & {\scriptsize{} \textbf{Margin type}} & {\scriptsize{} \textbf{Hard sample processin}g}\tabularnewline
				\hline
				{\scriptsize{} Softmax } & {\scriptsize{} -- } & {\scriptsize{} -- }   \tabularnewline
				\hline				
				{\scriptsize{} SphereFace \cite{17-SphereFace} } & {\scriptsize{} multiplicative angular margin} & {\scriptsize{} -- }  \tabularnewline 
				\hline
				{\scriptsize{} AM-Softmax \cite{18-AMSoftmax} } & {\scriptsize{} additive cosine margin } & {\scriptsize{} -- }  \tabularnewline
				
				\hline
				
				{\scriptsize{} ArcFace \cite{19-ArcFace}} & {\scriptsize{} additive angular margin } & {\scriptsize{} -- }  \tabularnewline		
				\hline
				
				\makecell[tl]{ \scriptsize{} F-AM-Softmax / \\ \scriptsize{} F-ArcFace \\ \scriptsize{} \cite{17-FocalLoss,19-MVSoftmax}  } & 	\multirow{3}{*}{\scriptsize{} additive cosine margin } & \makecell[tl]{\scriptsize{}   reduces the weight \\ \scriptsize{} of easier samples to \\ \scriptsize{} focus on hard samples. }   \tabularnewline
				
				\hline
				\makecell[tl] { \scriptsize{} HM-AM-Softmax/ \\ \scriptsize{} HM-ArcFace \\  \scriptsize{} \cite{16-HMSoftmax,19-MVSoftmax}} & 	\multirow{3}{*}{\scriptsize{} additive cosine margin } & \makecell[tl]{\scriptsize{}  indicates hard sample \\ \scriptsize{} discards easy sample.}   \tabularnewline
				
				\hline
				\makecell[tl] {\scriptsize{} MV-AM-Softmax/ \\ \scriptsize{} MV-ArcFace \cite{19-MVSoftmax}} & \makecell[tl]{\scriptsize{} combined cosine \\ \scriptsize{} and angular margin } & \makecell[tl]{\scriptsize{}  adjusts weight of \\ \scriptsize{}  mis-classified vectors.}   \tabularnewline
				
				\hline
				\multirow{2}{*}{\scriptsize{} CurricularFace \cite{20-CurricularFace}} & \multirow{2}{*}{\scriptsize{} additive angular margin } & \makecell[tl]{\scriptsize{} adaptively adjusts weight \\ \scriptsize{} via Curricular learning}   \tabularnewline				
				
				\hline
				\multirow{2}{*}{\scriptsize{} OTFace } & \makecell[tl]{\scriptsize{} additive cosine/ \\ \scriptsize{} angular margin } & \makecell[tl]{\scriptsize{} OT loss \\ \scriptsize{} from distribution scale }   \tabularnewline
				
				\hline
		\end{tabular}}{\small \par}
	\end{spacing}
\end{table}

\subsection{Comparisons with related methods}
In this subsection, we provide more analysis to compare the proposed OTFace with related Softmax-based methods. The Softmax-based methods aim to increase the decision margins for better feature embedding. SephereFace employs the multiplicative angular scheme to push the margin. Both AM-Softmax and CosFace apply the additive cosine margin to learn discriminative feature. In ArcFace, the additive angular margin scheme is used to increase the margin. 

Based on the above work, F-AM-Softmax employs the Focal loss to improve the performance of face verification. Focal loss focuses on the relatively hard samples by reducing the weights of easier samples. The drawback is that the definition of hard sample is not clear. For HM-AM-Softmax, the authors design the hard strategy to learn discriminative feature by using the high loss examples to construct the min-batches for training and the easy examples are ignored completely. MV-AM-Softmax explicitly considers the role of hard samples and concentrates on them to guide the discriminative feature learning by re-weighting the hard samples. For F-ArcFace, HM-ArcFace and MV-ArcFace, they just employ the ArcFace to replace AM-Softmax for learning discriminative feature. 
CurricularFace adaptively adjusts weight of hard samples in different training stages in conjunction with curricular learning.
Compared with previous work, OTFace borrows the idea of triplet to explore the hard sample groups and then combines the margin-based method and OT loss to enhance the discriminative power of embedding feature. Table \ref{tab:comparisons} lists the differences between OTFace and the related methods.

\section{Experiments}

In this section, we evaluated the proposed OTFace on eight published face databases: 
LFW \cite{Gary2007LFW}, AgeDB \cite{17AgeDB}, Cross-Age LFW (CALFW) \cite{17-CALFW}, Cross-Pose LFW (CPLFW) \cite{18-CPLFW}, CFP \cite{16CFP-FP}, IJB-B \cite{17-IJB-B}, IJB-C \cite{18-IJB-C} and MegaFace \cite{16-MegaFace}. 
We present the comparative results to demonstrate the advantages of the proposed methods over the state-of-the-art softmax-based methods for face recognition, such as AM-Softmax \cite{18-AMSoftmax}, ArcFace \cite{19-ArcFace}, F-AM-Softmax / F-ArcFace \cite{17-FocalLoss}, HM-AM-Softmax / HM-ArcFace \cite{16-HMSoftmax} and MV-AM-Softmax / MV-ArcFace \cite{19-MVSoftmax}. For these competed methods, we re-implement the codes of them and choose the best parameters according to their paper's suggestions. \textbf{Note that we use ResNet100 as backbone for AM-Softmax, F-AM-Softmax, HM-AM-Softmax, MV-AM-Softmax, ArcFace, F-ArcFace, HM-ArcFace and MV-ArcFace in our experiments}.

\subsection{Experimental Settings}
In our experiments, we employed MSIMV2 database as training set to conduct fair comparison with other comparative methods. MSIMV2 is a semi-automatic refined version of the MS-Celed-1M database and contains about 5,800,000 facial images from 85,000 persons. 
Similar with \cite{19-ArcFace}, we used the five facial points (two eyes, nose and two mouth corners) to align the faces and generate the cropped facial images ($112 \times 112$). The ResNet100 is used as the backbone network in our model. For sample strategy, we randomly sampled the images to construct the batches and set the batch size to 512. Then the SGD method was employed to optimize the training model. We set momentum to 0.9 and weight decay to 5e-4. The learning rate is set to 0.1 initially and the learning rate is divided by 10 when the epoch is 10, 18 or 22. The training stage was terminated when the epoch is 24. Here, four NVIDIA TITAN RTXs were used to complete the training processing step.

\subsection{Results on LFW}
LFW is the most widely used benchmark for unconstrained face verification and contains more than 13,000 images from 5749 persons collected from the web. These images have various facial expression, pose and illumination. The testing set is composed of 6,000 image pairs, which are separated into 10 folds and each fold contains 300 same and 300 non-same pairs. Here, we employed the aligned version of LFW. The images were simply cropped and normalized to 112$ \times $112 pixels.

In this experiment, we evaluated the efficiency of OTFaces and the comparative methods on LFW followed the standard test protocol. The VGGFace is the baseline in this experiment. The face verification rate of each method is listed in Table II. From Table II, we can see that most of the methods achieve satisfied results in this case. ArcFace and AM-Softmax perform slightly better than SphereFace, Marginal Loss and PFE. The performance of  MV-AM-Softmax is better than most competed methods since they focus on hard samples by adjusting the weights of samples in training stage. 
F-ArcFace and HM-ArcFace outperform F-AM-Softmax and HM-AM-Softmax, respectively. However, MV-ArcFace does not achieve the desired results.  The proposed OTFaces stand on top being better than other competed methods.

\subsection{Results on  AgeDB and CALFW }
In this section, we evaluate the performance of all methods for age-invariant face verification on AgeDB and CALFW databases. AgeDB is built by Imperial College London and contains 16,488 facial images from 568 famous persons. There are 29 facial images for each person and all images have a clean age label. The minimum and maximum age is 1 and 101, respectively. For AgeDB, the authors designed four test protocols to evaluate the effectiveness of the methods. 
For each protocol, AgeDB is divided into 10 folds and each fold is composed of 300 positive pairs and 300 negative pairs. 
The age gap of each face pair is fixed to 5, 10, 20 and 30 for different protocols. 
Here, we conducted experiments under the fourth protocol (age gap is 30) to evaluate the performance of each method. Table II lists the quantitative results of all methods on AgeDB. We can see that most OTFace(ArcFace) gives about 0.2\% improvement than the most competed methods and achieves the similar results as CurricularFace. 

CALFW is considered as a variant of LFW. Different from LFW, CALFW mainly employs the age property to re-organize the face image pairs. In CALFW, there are 4,025 persons and each person has two, three or four face images. Similar with LFW, CALFW provides 3,000 positive pairs and 3,000 negative pairs for face verification testing. For each pair, there is a significant age gap. Compared with AgeDB, CALFW is more challenged. We conducted experiments following the standard protocol. The face verification rates of all methods are listed in Table II. We can see clearly that the softmax-based methods are significantly better than VGGFace. The proposed OTFace(AM-Softmax) and OTFace(ArcFace) outperform the most methods. The result of MV-ArcFace is same as that of ArcFace. That means mis-classified based hard sample mining strategy has no improvements for ArcFace. However, it is beneficial to improve verification rates by employing the OT loss to describe the hard sample groups from the distribution scale.

\begin{table}
	\caption{\label{LFWs}Verification rates (\%) of each method on the LFW, AgeDB, CALFW, CPLFW and CFP Databases. "--" indicates that the authors did not report the results on the corresponding databases.}
	\begin{spacing}{0.9}
		\noindent \centering{}{\small{}}%
		\setlength{\tabcolsep}{2mm}{
			\begin{tabular}{l c c c c c}  
				\hline
				{\scriptsize{} Methods} & {\scriptsize{}LFW} & {\scriptsize{}AgeDB} & {\scriptsize{}CALFW}  
				& {\scriptsize{} CPLFW} & {\scriptsize{} CFP} \tabularnewline
				\hline
				{\scriptsize{} VGGFace \cite{15-VGGFace}} & {\scriptsize{}98.95} & {\scriptsize{} 85.1} & {\scriptsize{} 86.5}  & {\scriptsize{} --} & {\scriptsize{} --} \tabularnewline
				{\scriptsize{} Noisy Softmax \cite{17-NoisySoft}} & {\scriptsize{}99.18} & {\scriptsize{} --} & {\scriptsize{} 82.52} & {\scriptsize{} --} & {\scriptsize{} --}\tabularnewline
				
				{\scriptsize{} Marginal Loss \cite{17-MarginalLoss}} & {\scriptsize{}99.48} & {\scriptsize{} 95.7} & {\scriptsize{} --} & {\scriptsize{} --} & {\scriptsize{} --}\tabularnewline
				{\scriptsize{} SphereFace \cite{17-SphereFace}}  & {\scriptsize{}99.42} & {\scriptsize{} 97.16} & {\scriptsize{} 94.55} & {\scriptsize{} --} & {\scriptsize{} --} \tabularnewline
				
				{\scriptsize{} PFE \cite{19-PFE}}  & {\scriptsize{}99.70} & {\scriptsize{} --} & {\scriptsize{} --} & {\scriptsize{} --} & {\scriptsize{} 95.06} \tabularnewline
				
				{\scriptsize{} CurricularFace \cite{20-CurricularFace}}  & {\scriptsize{}99.80} & {\scriptsize{} \textbf{98.32}} & {\scriptsize{} \textbf{96.20}} & {\scriptsize{} 93.13} & {\scriptsize{} 98.37} \tabularnewline
				
				\hline
				{\scriptsize{} AM-Softmax } & {\scriptsize{}99.77} & {\scriptsize{} 98.03} & {\scriptsize{} 95.58} & {\scriptsize{} 92.53} & {\scriptsize{} 98.43}\tabularnewline
				{\scriptsize{} ArcFace } & {\scriptsize{}99.77} & {\scriptsize{} 98.00} & {\scriptsize{} 95.96} & {\scriptsize{} 93.05} & {\scriptsize{} 98.27} \tabularnewline
				{\scriptsize{} F-AM-Softmax } & {\scriptsize{}99.73} & {\scriptsize{} 98.13} & {\scriptsize{} 96.01} & {\scriptsize{} 92.80} & {\scriptsize{} 98.19} \tabularnewline
				{\scriptsize{} HM-AM-Softmax } & {\scriptsize{}99.76} & {\scriptsize{} 98.07} & {\scriptsize{} 96.01} & {\scriptsize{} 93.02} & {\scriptsize{} 98.43}\tabularnewline
				{\scriptsize{} MV-AM-Softmax } & {\scriptsize{}99.81} & {\scriptsize{} 98.15} & {\scriptsize{} 96.08} & {\scriptsize{} 93.08} & {\scriptsize{} 98.00} \tabularnewline
				
				\scriptsize{} F-ArcFace  & \scriptsize{}99.80 & \scriptsize{} 98.16  & \scriptsize{} 96.08 & \scriptsize{} 92.70  & \scriptsize{} 98.34 \tabularnewline
				
				\scriptsize{} HM-ArcFace  & \scriptsize{}99.81 & \scriptsize{} 98.18  & \scriptsize{} 96.13 & \scriptsize{} 92.90  & \scriptsize{} 98.33 \tabularnewline
				
				\scriptsize{} MV-ArcFace  & \scriptsize{}99.78 & \scriptsize{} 97.99  & \scriptsize{} 95.96 & \scriptsize{} 92.00 & \scriptsize{} 97.77 \tabularnewline
				
				\hline
				
				\scriptsize{} OTFace(AM-Softmax)  & \scriptsize{}99.82 & \scriptsize{} 98.15  & \scriptsize{} \textbf{96.18} & \scriptsize{} 93.20  & \scriptsize{} \textbf{98.64} \tabularnewline
				
				\scriptsize{} OTFace(ArcFace)  & \scriptsize{}\textbf{99.83} & \scriptsize{} \textbf{98.30}  & \scriptsize{} \textbf{96.18} & \scriptsize{} \textbf{93.62}  & \scriptsize{} 98.60 \tabularnewline
				\hline
				
		\end{tabular}}{\small \par}
	\end{spacing}
\end{table}

\subsection{Results on CPLFW and CFP}
In this section, we perform experiments on the CPLFW and the CFP databases to test all methods for pose-invariant face verification. CPLFW collected new facial images with large pose variations based on the identity list of LFW and each individual contains two facial images at least. Similar to LFW, there are ten folds and each fold contains 300 positive pairs and 300 negative pairs. In testing, CPLFW employs the same face verification protocols with LFW.
For CFP database, there are 7,000 facial images of 500 celebrities and each subject has 10 frontal and 4 profile face images. CFP also divides the images into 10 folds, each containing 350 positive and 350 negative pairs. Here, we mainly focus on the verification performance of all methods in dealing with the Frontal-Profile cases.

The verification rates of all methods on CPLFW and CFP are listed in Table II. The proposed OTFace(AM-Softmax) and OTFace(ArcFace) obviously beat the margin-based methods, including AM-Softmax and ArcFace. Compared with fusions of margin-based and mining-based methods (CurricularFace, MV-AM-Softmax/MV-ArcFace, HM-AM-Softmax/HM-ArcFace and F-AM-Softmax/F-ArcFace), OTFaces still perform better results than them. This is because OTFaces emplpoy ArcFace (or AM-Softmax) to maintain the performance of easy samples and uses OT loss to solve hard sample groups, leading results improvements. 
In fact, F-AM-Softmax, HM-AM-Softmax and MV-AM-Softmax also combine the hard sample mining strategy and AM-Softmax to improve the performance. However, the verification rates of HM-AM-Softmax have no improvements over AM-Softmax on the CFP. Both F-AM-Softmax and MV-AM-Softmax perform slightly poorer than AM-Softmax. Additionally, the results of MV-ArcFace, F-ArcFace and HM-ArcFace are poorer than ArcFace on the CPLFW. The possible reason is that the hard sample processing schemes of these methods do not fit to solve the pose-invariant face verification task.

\subsection{Results on IJB-B and IJB-C}
In this section, the proposed method is evaluated on the IJB-B and IJB-C databases, which are constructed based on IJB-A, a superset of IJB-A. IJB-B consists 1,845 subjects and each subject has two still images and one video at least. In total, there are 21,798 still images, and 55,026 video frames from the 7,011 videos. All bound boxes, eyes and noise local of faces were manually annotated. Actually, there are ten different protocols for testing face detection, identification, verification and clustering. Here, we mainly focus on the performance of all methods in accuracy of verification rate. In total, IJB-B provides 
10,270 genuine comparisons and 8,000,000 impostor comparisons. The IJB-C, which is considered as the extension of IJB-B and pays more attention to unconstrained media, which consists of 21,294 still images and 117,542 frames from 3,531 subjects. For IJB-C, we employed the mixed verification protocol to test all  methods in 1:1 verification. 

Table III tabulates the verification rates of all methods on the IJB-B and IJB-C. From Table III, we observe that OTFace(ArcFace) still performs better than all competed methods. Additionally, OTFaces again achieve the better results compared with the different types of method PFE. The performance of hard sample mining-based methods are better than that of AM-Softmax.  However, ArcFace achieves the similar results to MV-AM-Softmax and outperforms the F-AM-Softmax and HM-AM-Softmax. The possible reason is that additive angular margin loss is better than that of additive cosine margin in this case.
Fig.5 shows the ROC curves of OTFaces and the comparable methods on IJB-B and IJB-C. From Fig. 5, our method still achieves the leading results in most cases.

\begin{table}
	\caption{\label{IJB-B/C}Verification rates (FAR=$ 1e-4 $) of each method on the IJB-B and IJB-C databases. "--" indicates that the authors did not report the results on the corresponding databases. }
	\begin{spacing}{0.9}
		\noindent \centering{}{\small{}}%
		\setlength{\tabcolsep}{1.5mm}{
			\begin{tabular}{l c c }
				\hline
				{\scriptsize{}Methods} & {\scriptsize{}IJB-B} & {\scriptsize{}IJB-C}  \tabularnewline
				\hline
				{\scriptsize{}PFE \cite{19-PFE}} & {\scriptsize{}--} & {\scriptsize{}93.25} \tabularnewline
				{\scriptsize{}AdaCos \cite{19-AdaCos} } & {\scriptsize{}--} & {\scriptsize{}92.4}  \tabularnewline
				{\scriptsize{}CurricularFace \cite{20-CurricularFace} } & {\scriptsize{}94.8} & {\scriptsize{}96.1}  \tabularnewline
				
				\hline
				{\scriptsize{}AM-Softmax } & {\scriptsize{}93.54} & {\scriptsize{}95.02} \tabularnewline
				{\scriptsize{}ArcFace } & {\scriptsize{}94.26} & {\scriptsize{}95.73} \tabularnewline
				{\scriptsize{}F-AM-Softmax } & {\scriptsize{}93.60} & {\scriptsize{}95.19}  \tabularnewline
				{\scriptsize{}HM-AM-Softmax } & {\scriptsize{}93.80} & {\scriptsize{}95.20}  \tabularnewline
				{\scriptsize{}MV-AM-Softmax } & {\scriptsize{}94.08} & {\scriptsize{}95.57}  \tabularnewline
				
				{\scriptsize{}F-ArcFace } & {\scriptsize{}94.43} & {\scriptsize{}95.84}  \tabularnewline
				{\scriptsize{}HM-ArcFace } & {\scriptsize{}94.32} & {\scriptsize{}95.73}  \tabularnewline
				{\scriptsize{}MV-ArcFace } & {\scriptsize{}94.01} & {\scriptsize{}95.59}  \tabularnewline
				
				\hline
				{\scriptsize{}OTFace(AM-Softmax) }  & { \scriptsize{}94.25 } & { \scriptsize{}95.70 }   \tabularnewline
				
				{\scriptsize{}OTFace(ArcFace) } & {\scriptsize{} \textbf{94.84} } & {\scriptsize{} \textbf{96.15} }  \tabularnewline
				
				\hline
		\end{tabular}}{\small \par}
	\end{spacing}
\end{table}

\begin{figure}
	\noindent \begin{centering}
		\includegraphics[scale=0.3]{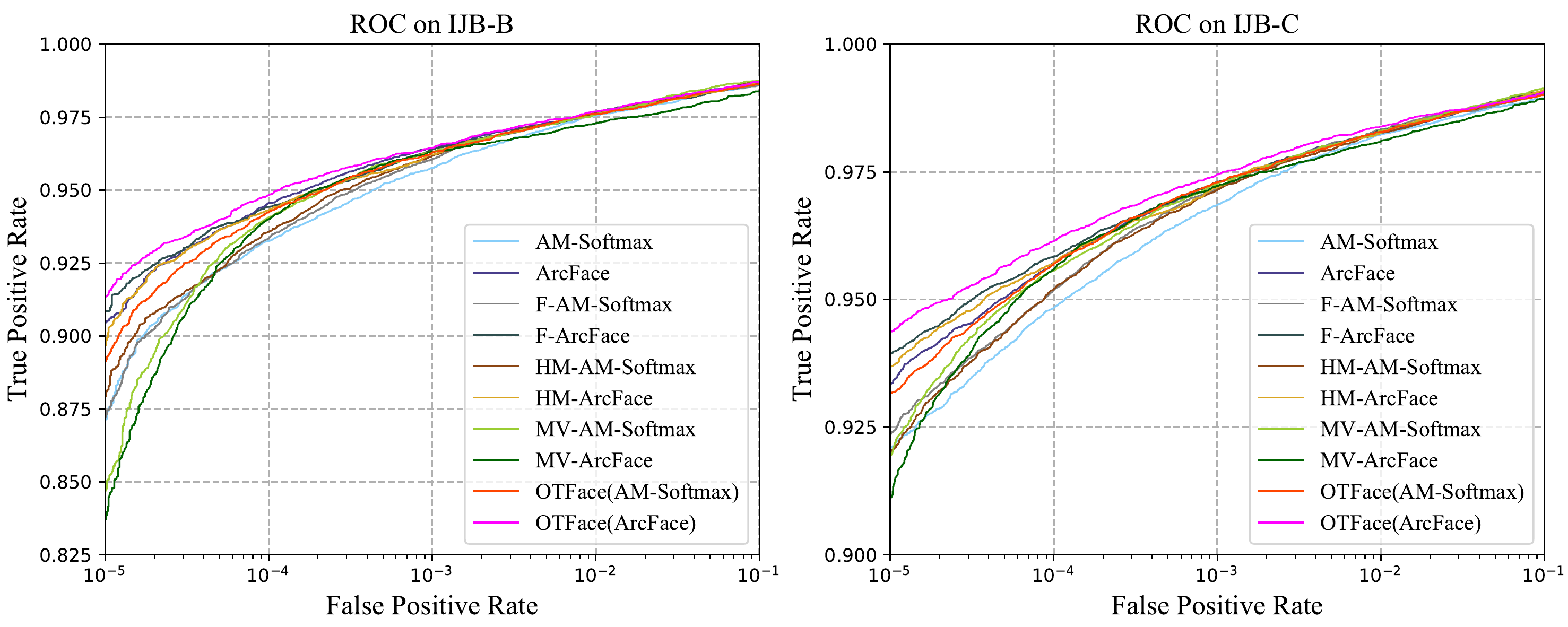}
		\par\end{centering}
	\begin{singlespace}
		\caption{ROC curves of each method on IJB-B and IJB-C. }
	\end{singlespace}
\end{figure} \label{fig:IJB-B/C}


\subsection{Results on MegaFace}
In this section, we conduct experiments on the MegaFace database, which is a very large scale database. The MegaFace dataset includes one million images of more than 690,000 individuals as the gallery set. FaceScrub \cite{14-FaceScrub} and FG-NET were used as the probe set. FaceScrub includes 100K images from 530 celebrities and has a large variation across images of the same person. On MegaFace, both identification and verification results were provided in our experiments under the protocols with large training set (contains about 5M images). For face verification, there are about  0.2 million positive pairs and 4 billion negative pairs, which make the task more challenging. 

Here, we mainly employ the MSIMV2 to train our model. The face identification results for rank 1 of all methods and the face verification rates at FAR=$ 1e-6 $ of all methods are listed in Table IV. From Table IV we can see that our OTFace(ArcFace) achieves the top results for both identification (98.52$ \% $) and verification (98.82$ \% $). OTFace(AM-Softmax) also beats the competed methods and performs slightly poor than OTFace(ArcFace). This further validates the advantages of OTFace to solve hard sample groups by using OT loss. It is known that MV-AM-Softmax, HM-AM-Softmax and F-AM-Softmax also employ the hard sample mining strategy in conjunction with AM-Softmax to learn discriminative feature. However, these methods did not show the performance improvements over AM-Softmax since the hard sample strategies of these methods are not suitable to improve the recognition performance when has a large variation across images of the same class. MV-ArcFace and HM-ArcFace also perform slightly poor than ArcFace, which is consistent with above explanation.
In summary, OTFaces achieve the best results among all methods under various face verification cases. 

\begin{table}
	\caption{\label{tab:MegaFace}The identification results for Rank 1 and the verification rates of each method on the MegaFace database. The verification rates are computed at FAR=$ 1e-6 $.}
	\begin{spacing}{0.9}
		\noindent \centering{}{\small{}}%
		\setlength{\tabcolsep}{4mm}{
			\begin{tabular}{l c c c}
				\hline
				{\scriptsize{}Methods}  &  {\scriptsize{}Identification} & {\scriptsize{}Verification}  \tabularnewline			
				\hline
				
				{\scriptsize{}PFE \cite{19-PFE} } & {\scriptsize{}72.43} & {\scriptsize{}92.93}  \tabularnewline
				
				{\scriptsize{}AdaCos \cite{19-AdaCos} }  & {\scriptsize{}97.41} & {\scriptsize{}--}  \tabularnewline
				
				{\scriptsize{}CurricularFace \cite{20-CurricularFace} } & {\scriptsize{}98.25} & {\scriptsize{}98.44}  \tabularnewline
				
				\hline
				{\scriptsize{}AM-Softmax }  & {\scriptsize{}98.20} & {\scriptsize{}98.50}  \tabularnewline
				{\scriptsize{}ArcFace }  & {\scriptsize{}98.34} & {\scriptsize{}98.55} \tabularnewline
				{\scriptsize{}F-AM-Softmax }  & {\scriptsize{}98.09} & {\scriptsize{}98.60}  \tabularnewline
				{\scriptsize{}HM-AM-Softmax }  & {\scriptsize{}98.07} & {\scriptsize{}98.52}  \tabularnewline
				{\scriptsize{}MV-AM-Softmax }  & \scriptsize{}97.55 & \scriptsize{}98.13\tabularnewline
				
				{\scriptsize{}F-ArcFace } & {\scriptsize{}98.37} & {\scriptsize{}98.60}  \tabularnewline
				{\scriptsize{}HM-ArcFace } & {\scriptsize{}98.27} & {\scriptsize{}98.55}  \tabularnewline
				{\scriptsize{}MV-ArcFace } & {\scriptsize{}98.22} & {\scriptsize{}98.28}  \tabularnewline
				\hline
				
				{\scriptsize{}OTFace(AM-Softmax) }  & \scriptsize{}98.49 & \scriptsize{}98.81\tabularnewline
				
				{\scriptsize{}OTFace(ArcFace) }  & \scriptsize{}\textbf{98.52}  & \scriptsize{}\textbf{98.82} \tabularnewline
				\hline
		\end{tabular}}{\small \par}
	\end{spacing}
\end{table}

\begin{table*}
	\caption{\label{tab:Ablation-2}The results (\%) of OTFace with different scale of feature map on the eight benchmark databases. }
	\begin{spacing}{0.9}
		\noindent \centering{}{\small{}}%
		\setlength{\tabcolsep}{2mm}{
			\begin{tabular}{l c c c c c c c c c}
				\hline
				\multirow{2}{*}{\scriptsize{}Methods (feature map size, feature map Number)}  &  \multirow{2}{*}{\scriptsize{}LFW} &  \multirow{2}{*}{\scriptsize{}AgeDB} 
				&  \multirow{2}{*}{\scriptsize{}CALFW} &  \multirow{2}{*}{\scriptsize{}CPLFW} &  \multirow{2}{*}{\scriptsize{}CFP}
				& \multirow{2}{*}{\scriptsize{}IJB-B} & 	\multirow{2}{*}{\scriptsize{}IJB-C} &
				\multicolumn{2}{c }{\scriptsize{}MegaFace} \tabularnewline		
				
				\cline{9-10}
				& {\scriptsize{}\space} & {\scriptsize{}\space} & {\scriptsize{}\space} & {\scriptsize{}\space} & {\scriptsize{}\space} & {\scriptsize{}\space} & {\scriptsize{}\space} & {\scriptsize{}Identification} & {\scriptsize{}Verification}  \tabularnewline	
				
				\hline
				{\scriptsize{}OTFace(AM-Softmax) ($ 56 \times 56, 64 $) }  & {\scriptsize{} 99.80} & {\scriptsize{} 98.15} & {\scriptsize{} 95.87} & {\scriptsize{} 92.85} & {\scriptsize{} 98.53} 
				& {\scriptsize{} 94.22} & {\scriptsize{}95.59}  & {\scriptsize{}98.24} & {\scriptsize{}98.61}\tabularnewline
				
				{\scriptsize{}OTFace(AM-Softmax) ($ 28 \times 28, 128 $) }  & {\scriptsize{}\textbf{ 99.82}} & {\scriptsize{} 98.10} & {\scriptsize{} 96.03} & {\scriptsize{} \textbf{93.20}} & {\scriptsize{} \textbf{98.64}} 
				& {\scriptsize{} 94.00} & {\scriptsize{}95.59}  & {\scriptsize{}\textbf{98.49}} & {\scriptsize{}\textbf{98.81}}\tabularnewline
				
				{\scriptsize{}OTFace(AM-Softmax)($ 14 \times 14, 256 $) } & {\scriptsize{} 99.80} & {\scriptsize{} \textbf{98.15}} & {\scriptsize{} \textbf{96.18}} & {\scriptsize{} 92.96} & {\scriptsize{} 98.54}
				& {\scriptsize{} \textbf{94.25}} & {\scriptsize{}\textbf{95.73}} & {\scriptsize{}98.36} & {\scriptsize{}98.75} \tabularnewline
				{\scriptsize{}OTFace(AM-Softmax) ($ 7 \times 7, 512 $)}  & {\scriptsize{} 99.77} & {\scriptsize{} 98.15} & {\scriptsize{} 96.03} & {\scriptsize{} 93.11} & {\scriptsize{} 98.56} 
				& {\scriptsize{} 94.04} & {\scriptsize{}95.50} & {\scriptsize{} 98.38 } & {\scriptsize{}98.71} \tabularnewline
				\hline
				
				{\scriptsize{}OTFace(ArcFace) ($ 56 \times 56, 64 $) }  & {\scriptsize{} 99.78} & {\scriptsize{} 98.07} & {\scriptsize{} 96.03} & {\scriptsize{} 92.95} & {\scriptsize{} 98.47} 
				& {\scriptsize{} \textbf{94.84}} & {\scriptsize{}\textbf{96.15}}  & {\scriptsize{}98.46} & {\scriptsize{}98.72}\tabularnewline
				
				{\scriptsize{}OTFace(ArcFace) ($ 28 \times 28, 128 $) }  & {\scriptsize{} \textbf{99.83}} & {\scriptsize{} 98.18} & {\scriptsize{} 96.10} & {\scriptsize{} \textbf{93.62}} & {\scriptsize{} \textbf{98.60}} 
				& {\scriptsize{} 94.70} & {\scriptsize{}96.07}  & {\scriptsize{}\textbf{98.52}} & {\scriptsize{}\textbf{98.82}}\tabularnewline
				
				{\scriptsize{}OTFace(ArcFace) ($ 14 \times 14, 256 $) } & {\scriptsize{} 99.80} & {\scriptsize{} \textbf{98.30}} & {\scriptsize{} \textbf{96.18}} & {\scriptsize{} 92.90} & {\scriptsize{} 98.37}
				& {\scriptsize{} 94.65} & {\scriptsize{}96.04} & {\scriptsize{}98.50} & {\scriptsize{}98.78} \tabularnewline
				{\scriptsize{}OTFace(ArcFace) ($ 7 \times 7, 512 $)}  & {\scriptsize{} 99.82} & {\scriptsize{} 98.22} & {\scriptsize{} 96.10} & {\scriptsize{} 93.35} & {\scriptsize{} 98.51} 
				& {\scriptsize{} 94.73} & {\scriptsize{}96.09} & {\scriptsize{} 98.51 } & {\scriptsize{}98.76} \tabularnewline
				
				\hline
		\end{tabular}}{\small \par}
	\end{spacing}
\end{table*}

\subsection{Further Discussion}
In this section, we firstly investigate the different configurations of the proposed OTFace based on the same test protocols on eight benchmark databases. 
In fact, OTFace can be considered as an enhanced version of ArcFace (or AM-Softmax) since OTFace is degraded to ArcFace (or AM-Softmax) when we remove the hard sample mining scheme with OT loss. In previous experiments (Tables 2, 3 and 4), we can see clearly that the proposed OTFace performs better results than ArcFace (or AM-Softmax)  in various face verification cases. These experiments show that OT loss in conjunction with hard sample group mining scheme brings performance improvements. 

In OTFace, the feature maps of the high level convolutional layer are used to compute the OT loss. Here, we evaluate the performance of OTFace(AM-Softmax) and OTFace(ArcFace) by using feature maps with different granularity. Table V tabulates the results of OTFace(AM-Softmax) and OTFace(ArcFace) with different sizes of feature maps. From Table V, we find that both OTFace(AM-Softmax) and OTFace(ArcFace) perform better results in most cases when the feature map size is 28 $ \times $ 28. It is known that the larger feature map contains more geometry distribution information. However, the smaller feature map contains more semantic information.
The possible reason is that the combination of geometry distribution information and semantic information is near optimal.
The results of OTFaces with 14$ \times $14 feature maps are better than other cases for age-invariant face verification on AgeDB and CALFW.
Because age invariant face representation requires more semantic information with suitable geometry information.
Many face images of IJB-B and IJB-C are blurry, OTFace achieves the best results by using larger feature map.
The probable reason may be that it is difficult to capture the semantic information of low-quality face image, however, the geometry distribution information plays more important role in this case.


\begin{figure*}
	\noindent \begin{centering}
		\includegraphics[scale=0.58]{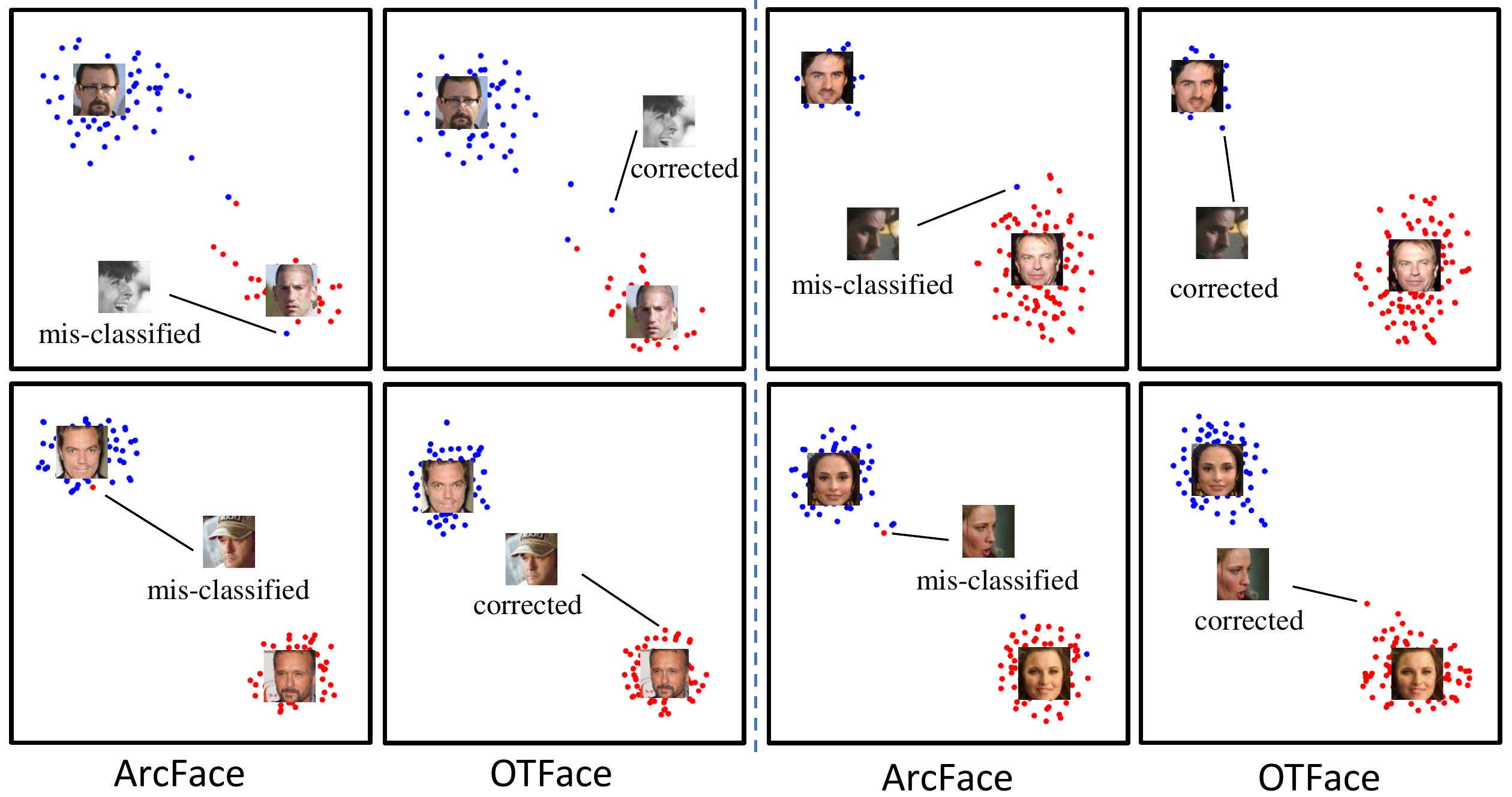}
		\par\end{centering}
	\caption{\small The t-SNE visualization of the learned features by ArcFace and OTFace. Some examples show that the hard samples are misclassified by using ArcFace while OTFace can give the corrected results.}
\end{figure*} \label{fig:visual}

Here, we also provide the t-SNE visualization of the learned feature by the ArcFace and OTFace to show the distribution changed in the hard samples before and after applying the OT loss as shown in Fig.6.
From Fig.6, we can see clearly that the feature distribution using OT loss is more compact than that of ArcFace. The hard sample is misclassified by using ArcFace. However, OTFace can give corrected results. This further demonstrates the advantages of our OT loss. 

\section{Conclusion}
This paper has developed a deep face representation model OTFace that combines the OT loss and margin-based Softmax to formulate the dual shot loss scheme. Specifically, we employ OT loss from the distribution scale to characterize the hard sample groups, which are mined based on the idea of triplet loss. Margin-based softmax (AM-Softmax or ArcFace) is actually used to maintain the performance of easy samples. In this way, the proposed OTFace can guide the deep CNN to learn the compact and discriminative feature. Experimental results on various face databases also verify the effectiveness of OT loss to solve hard samples, resulting in further performance improvements. However, it is difficult to enhance the feature discrimination by OT loss in dealing with the low-resolution face images. 
The possible reason is that the distribution property cannot reveal the similarity well between the low-resolution image and the corresponding high-resolution one. 
In future work, we will explore the local structure of OT to further improve the robustness of OT loss. It is also interesting in investigating more metric schemes to reduce the gap between low-resolution and high-resolution images efficiently.


%





\ifCLASSOPTIONcaptionsoff
  \newpage
\fi



%

\bibliographystyle{IEEEtran}
\bibliography{My_references}

%
%

%








\end{document}